# Improved underwater image enhancement algorithms based on partial differential equations (PDEs)

## U. A. Nnolim

*Department of Electronic Engineering, University of Nigeria Nsukka, Enugu, Nigeria*

## Abstract

*The experimental results of improved underwater image enhancement algorithms based on partial differential equations (PDEs) are presented in this report. This second work extends the study of previous work and incorporating several improvements into the revised algorithm. Experiments show the evidence of the improvements when compared to previously proposed approaches and other conventional algorithms found in the literature.*

### 1. Introduction

Numerous works involving underwater image enhancement have been published in the literature [1] [2] [3] [4] [5] [6] [7] [8] [9] [10] [11] [12] [13] [14] [15] [16] [17] [18] [19] [20] [21] [22] [23] [24] [25] [26] [27] with newer techniques adding to the continuously expanding body of work. These algorithms span a wide array of methods and approaches. However, it is difficult to observe extensive testing of several well-known images using several of the developed algorithms. Additionally, it is a challenge to obtain all the implementations of the published works from the authors for evaluation and testing with additional image data.

Previous work [28] involved the development of several algorithms based on partial differential equations for underwater image enhancement based on modifications and combinations of natural image enhancement algorithms [29] [30] such as the contrast limited adaptive histogram equalization (CLAHE) [31]. The results showed several advantages and improvements of the proposed algorithms over previous works from the literature [29] [30]. Additionally, the PDE-based framework enabled greater control over the various combined processes [32]. This was impressive considering that single image-based enhancement methods do not operate with any image formation or acquisition data [1]. Thus, it would not be unexpected if such algorithms were to fail or be wholly inadequate for processing underwater images, which suffer from unique problems due to the aquatic environment [1].

However, the proposed algorithms were unable to process adequately a few images, which had unique problems that could not be resolved completely with the earlier proposed approaches. This led to the evaluation of the key algorithms that were consistently effective and combining them with additional colour correction methods suited to adequately processing the affected images [29] [30].

The results of these improvements are presented for comparison with previous approaches and other algorithms from established works found in the literature.

Motivation for improving on this work was partially due to the encouraging results and confirmation from other authors who have mentioned the usage of conventional algorithms in underwater image enhancement [5] [10] [17] [22]. On the other hand, improving the results of the previous approaches presents an opportunity for further study.

### 2. Background on PDE formulation for image enhancement

The initial framework for the proposed algorithms is rooted in earlier works by Shapiro, et and the basic formulation [33] [32] was given as;

$$\frac{\partial I(x,y,t)}{\partial t} = \lambda D(\|\nabla I(x,y,t)\|)\text{div}\left(\frac{\nabla I(x,y,t)}{\|\nabla I(x,y,t)\|}\right) + [f\{I(x,y,t)\} - I(x,y,t)] \qquad (1)$$

We modified this approach in previous work [29] [30] to yield the modified and updated models;

$$\frac{\partial I(x,y,t)}{\partial t} = \lambda D(\|\nabla I(x,y,t)\|)\text{div}\left(\frac{\nabla I(x,y,t)}{\|\nabla I(x,y,t)\|}\right) + f\{I(x,y,t)\} + C(x,y,t) \qquad (2)$$

$$\frac{\partial I(x,y,t)}{\partial t} = \lambda D(\|\nabla I(x,y,t)\|)\text{div}\left(\frac{\nabla I(x,y,t)}{\|\nabla I(x,y,t)\|}\right) + [f_l\{I(x,y,t)\} - I(x,y,t)] + f_g\{I(x,y,t)\} + [C(x,y,t)] \qquad (3)$$

In (1), (2) and (3), $I(x,y,t)$ is the continuous image, with coordinates in (horizontal and vertical) spatial and temporal coordinates, x, y and t respectively. The smoothing term, $\lambda D(\|\nabla I(x,y,t)\|)\text{div}\left(\frac{\nabla I(x,y,t)}{\|\nabla I(x,y,t)\|}\right)$ is the anisotropic diffusion term with control parameter, $\lambda > 0$, $\nabla$ is the gradient operator, $\| \quad \|$ as the norm and div as the divergent operator. The local and global contrast operators, $f_l$, $f_g$ in addition to the colour correction term, $C(x, y, t)$ are defined by contrast enhancement and colour correction functions. The functions used are given as; $f_l\{I(x, y, t)\} = CLAHE\{I(x, y, t)\}$ and $f_g\{I(x, y, t)\} = GOC2\{I(x, y, t)\}$, though any global contrast enhancement operator could be used. For (2), we use the term $C(x, y, t) = \frac{(I(x, y, t) - m)}{\sigma}$ to ensure a gradual evolution using mode, $m$, in the absence of a fidelity term while for (3) we use $C(x, y, t) = \frac{(I(x, y, t) - \mu)}{\sigma}$ with mean, $\mu$, to speed up the process since the presence of the fidelity term in the contrast enhancement in the latter ensures stability and convergence.

We continue to build on this approach by cascading certain various algorithms to ensure that local and global image features are adequately enhanced in the process. This further boosts the strengths of the individual algorithms, minimizing their disadvantages while amplifying their strengths. For example, the CLAHE method is a powerful localized operator when applied correctly. However, colour distortion is unavoidable with such algorithms. Thus, globalized operators help to remedy this problem and the sequence in which each of these operators are applied depends on the image as will be seen in experiments [29] [30] [34].

### 3. Improved PDE-based algorithm
Based on experiments, mathematical derivation and proof, we select the PWL algorithm since it is a generalization of the various contrast stretching approaches. Thus, based on findings from this work [34] and previous study [29] [30], we combine the local and global operations in cascaded form such as; $f_l\{f_g\{I(x, y, t)\}\} = f_{lg}\{I(x, y, t)\} = CLAHE(PWL(\{I(x, y, t)\}))$ and introduce additional control parameters to further regularize and regulate the various processes within the new PDE [34]. The improved PDE-based approaches are discussed in this section and utilize additional components for processing specific images. The results of the new approach [34] and the previous work [29] [30] were subsequently compared.

### 3.1 Comparison of selected PDE-based approaches
The results of using the various selected contrast enhancement and colour correction algorithms are presented in this subsection. These include the key selected algorithms from previous work incorporated into a PDE-based framework [29] [30]. We present the images used in the various experiments in Fig. 1.

In previous work, we performed numerous experiments with these images using a number of conventional image enhancement algorithms to obtain suitable candidates for augmentation within a PDE-based framework [29] [30]. In this work, we now further explore the modified approaches. It was observed that some images yielded better results using a global-local process than a local-global one [34]. Thus we show results using alternative sequences of global and local contrast operators and based on results, certain images are favoured by a particular sequence of operations over the other [34].

Based on the results in Fig. 2, we obtain fairly consistent performance indicating that these algorithms are effective for most images. We focus on trying to improve results for the few images for which the approaches perform poorly.

### 3.2 Further improvements and additions to the proposed approaches
Not all images are adequately enhanced by the various substitutions of the selected contrast enhancement and colour correction operators. For example, in Fig. 2, the images used in works by Bianco, et al [6] (based on the work by Reinhard, et al [35]) and Gouinaud, et al [16] show poor colour correction results. Based on the image histogram analysis in previous work [29] and shown in Fig. 3, it is observed that the red, green and blue channels are not in alignment with each other. Thus, it would be difficult to realign the colours with RGB-based colour correction schemes. Thus, we explore the incorporation of the methods used in Gouinaud and Bianco's work into the proposed algorithms. The proposed modifications are shown in Fig. 4. For easy notation, we denote the additional approaches as PA-1 and PA-2 as shown in Fig. 4.

Based on investigations, the main component for the colour correction for such images is the RGB2XYZ operation, which can be performed prior or after enhancement using the proposed PDE-based PWL-CLAHE scheme. The greyed out boxes in Fig. 4(b) are for the Fuzzy Homomorphic Enhancement [36] [37] and Piece-wise linear transform-based (PWL) [38] enhancement components, which are optional. They are normally used when the output image is dark or faded to restore contrast to the processed image.

Based on Fig. 4, PA-1 is relatively more complicated than PA-2 and though both algorithms clearly eliminate the colour cast, each is better suited to a particular image than the other. Utilizing perceptual colour spaces such as HSI or HSV fails in this case to yield pleasing results for these particular images. For example, using Iqbal's scheme in this case yielded no effect on the colour cast effects. This is easily understood from the histogram analysis as discussed in previous work [29].

## 4 Experiments and results

Further experiments are performed to verify the efficacy of incorporating these algorithms into the proposed framework. The results are shown in Fig. 5 and 6.

Based on the visual results in Fig. 5, we can see that the hybrid approach yields much better colour correction and contrast enhancement results. The image results from Bianco still appear faded and hazy compared to the results using the proposed hybrid approach. However, for the results by Gouinaud, et al [16], the results are darker using the proposed modifications. Thus, in order to refine the process, we devise a new scheme to process these images in order to avoid the complexity of the Colour Logarithmic Image Processing (CoLIP) approach used in work by Gouinaud et al [16].

The results obtained using the second scheme (PA-2) are shown in Fig. 6 and compared with the results using Gouinaud, et al [16]. It is clearly seen that there is still a considerable amount of blue haze in the results obtained using the CoLIP approach. In comparison, the modified approach yields images with a larger portion of the dominant blue hue removed. However, there is colour distortion in addition to contrast enhancement compared with the CoLIP method. Also, some colour distortion is observed with the images used by Bianco et al [6] though there is a great deal of colour cast removal (first row of Fig. 6). Nevertheless, the additional proposed colour correction scheme of PA-2 is much simpler than the CoLIP.

Ultimately, we have devised solutions to adequately process the outlier images in addition to the initial scheme that works well for most images. Thus, the proposed approaches show much improved results, though there are still other issues to be resolved for these type of underwater images. Additional future work may involve the possibility of devising a scheme to classify underwater images based on certain unique features. However, it will be not an easy task to numerically quantify such a subjective attribute.

## Conclusion and future work

We have presented an improved PDE-based scheme for single underwater image enhancement in addition to colour correction pre- and processing steps based on colour space conversions. The proposed schemes are stable and effective for most images in addition to yielding better results than most algorithms from the literature. Additional improvements are also devised and proposed for handling the rare cases of failure of the proposed approaches. The key idea behind the schemes is the use of modified algorithms in the absence of image information to enhance underwater images. Thus, the approach is quite effective and some evidence in the literature validates the fundamental ideas used in developing the proposed techniques. Future work would involve automated recognition and classification of such outlier images and improving results for the specific cases in terms of colour enhancement, where possible.

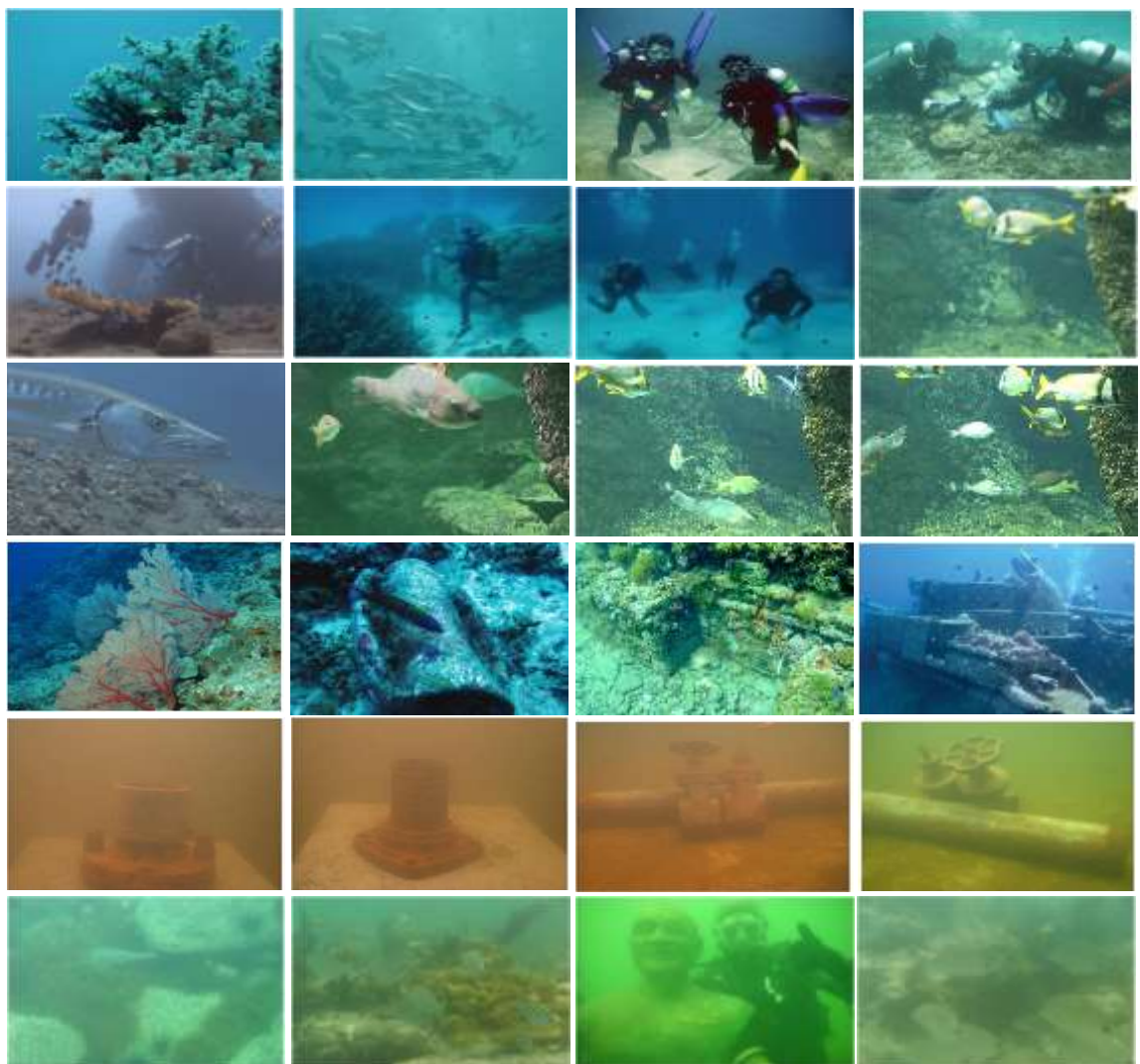

(a)

| (a) sea plants | (b) fishes1 | (c) divers1 | (d) divers2 |
|---|---|---|---|
| (e) divers3 | (f) diver | (g) divers4 | (h) fishes2 |
| (i) barracuda | (j) fishes3 | (k) fishes4 | (l) fishes5 |
| (m) seaplants2 | (n) ocean jar | (o) ocean floor | (p) ship wreck |
| (q) object1 | (r) object2 | (s) object3 | (t) object4 |
| (u) various | (v) fishes6 | (w) statue | (x) fishes7 |

**KEY**

(b)

**Fig. 1** (a) Underwater images used in experiments (b) key to figures

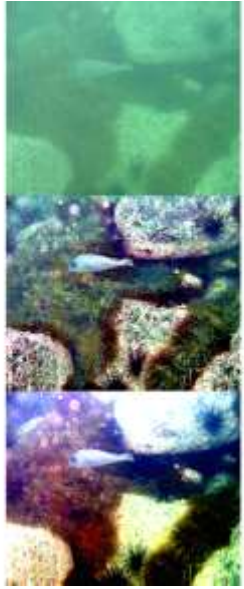
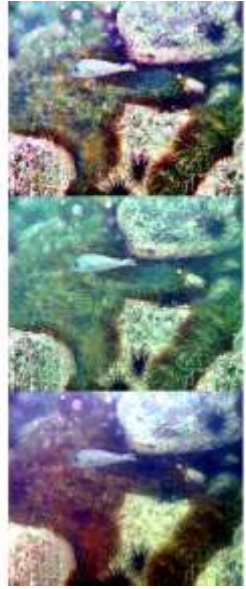
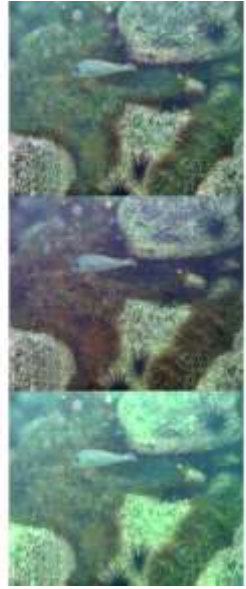
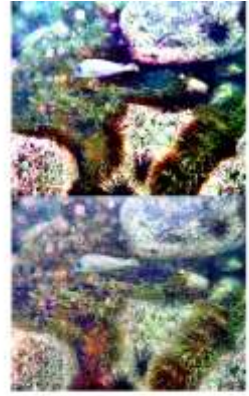

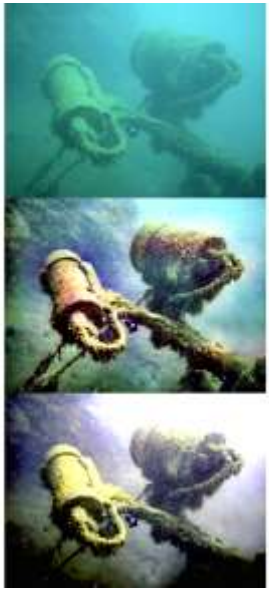
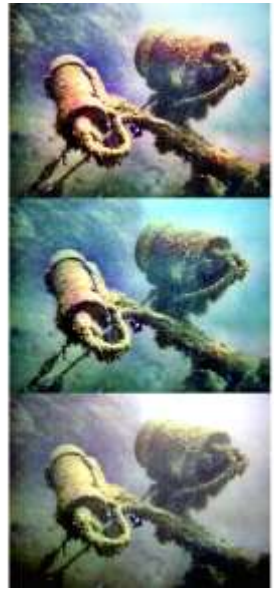
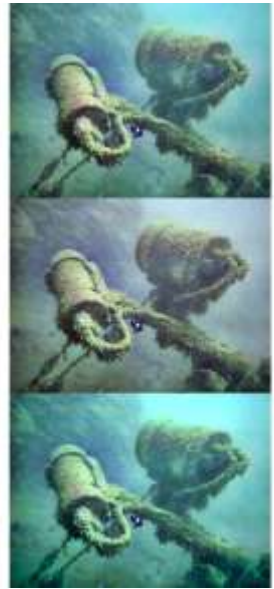
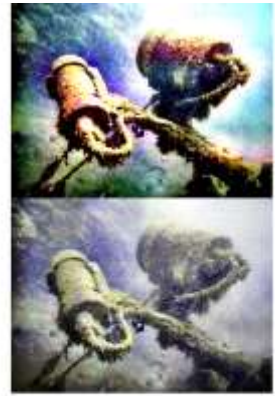

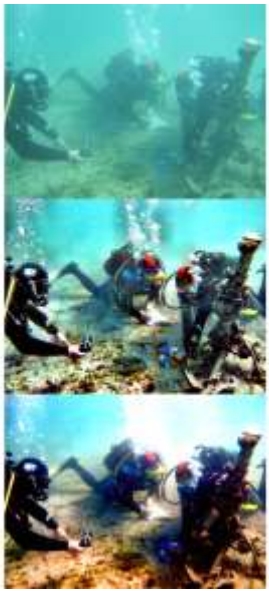
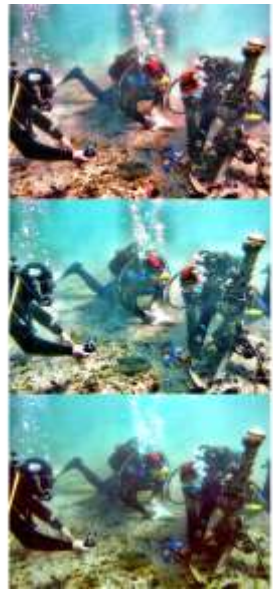
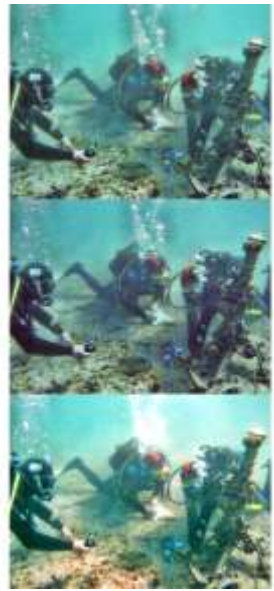
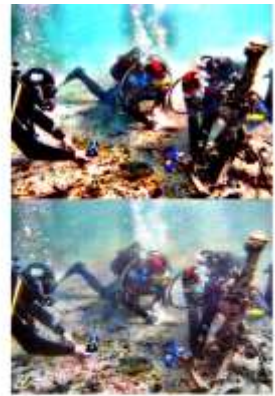

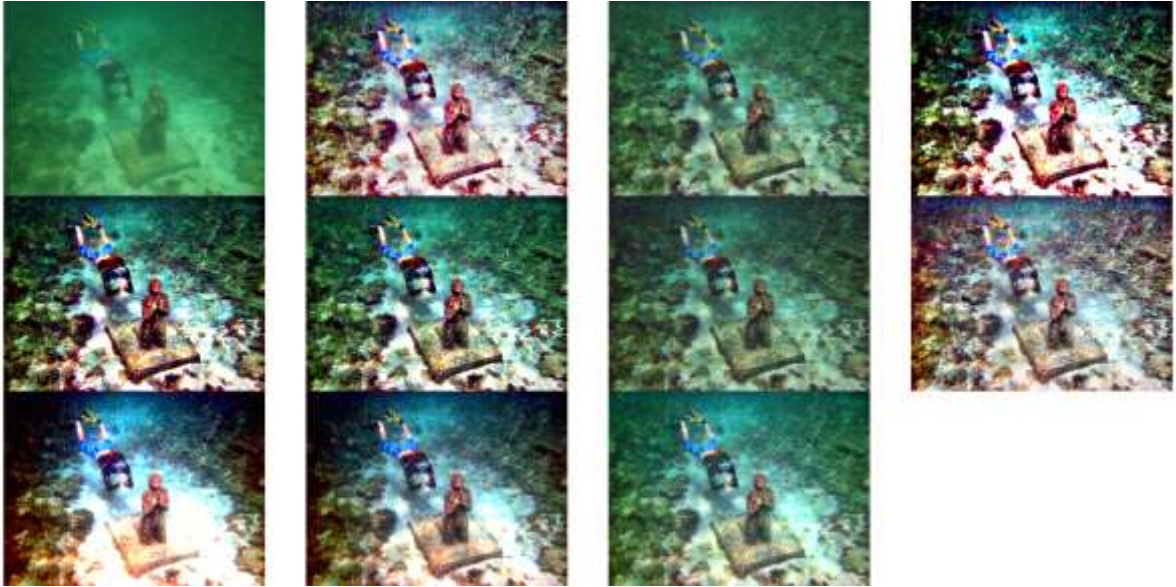

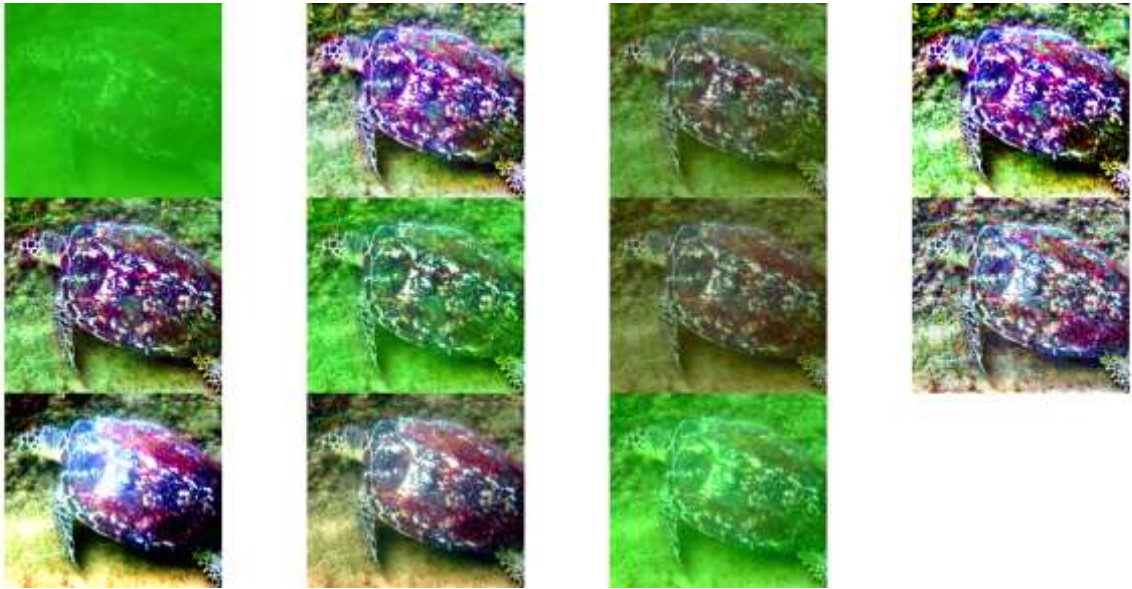

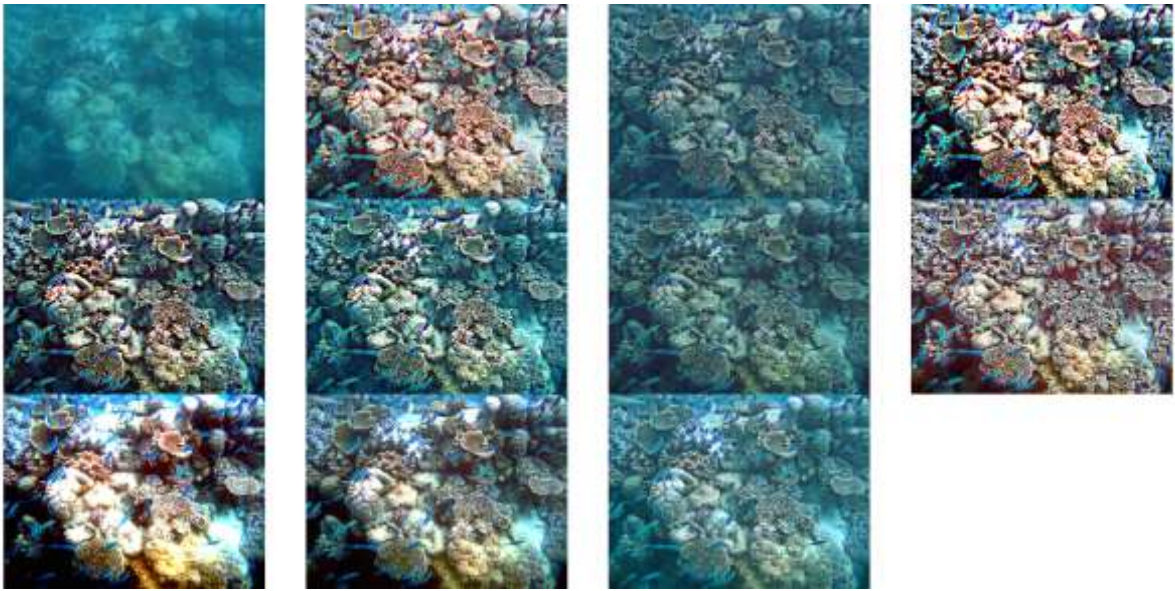

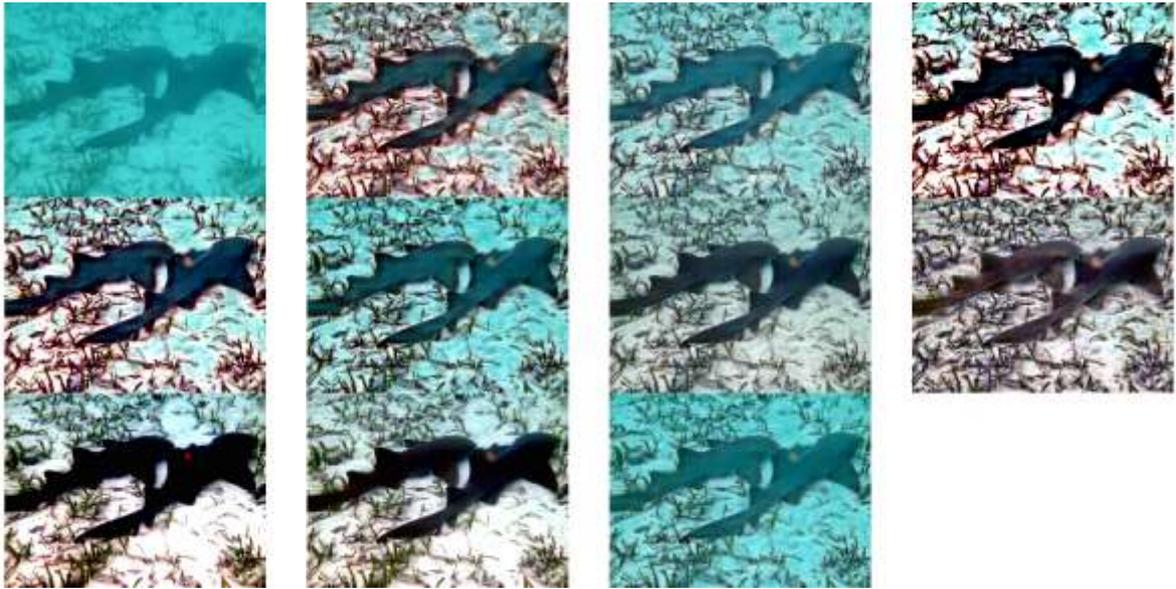

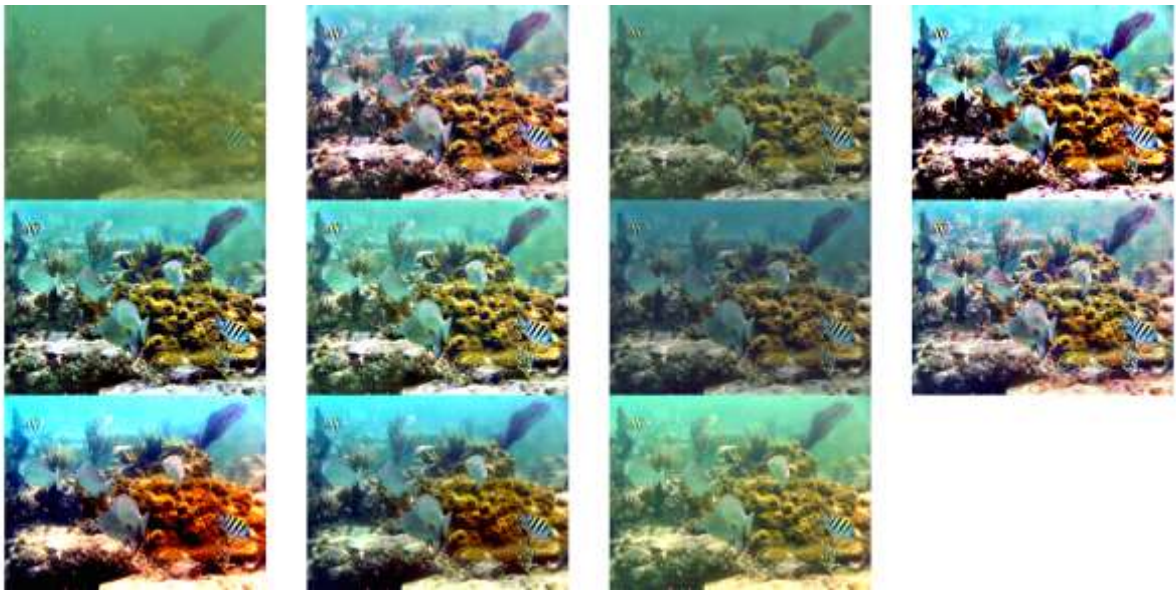

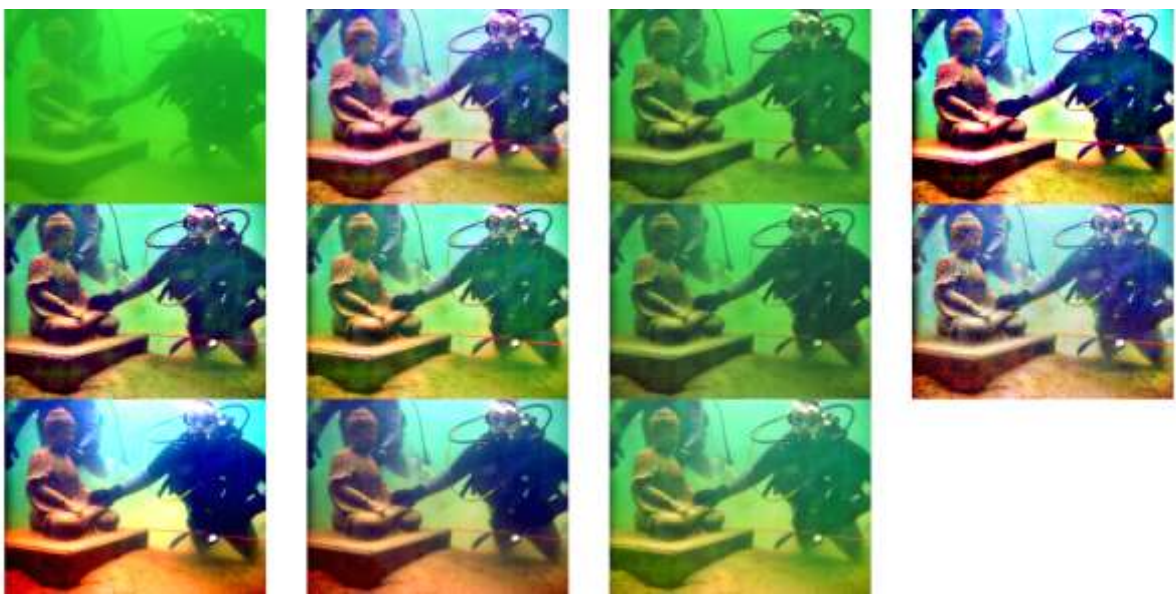

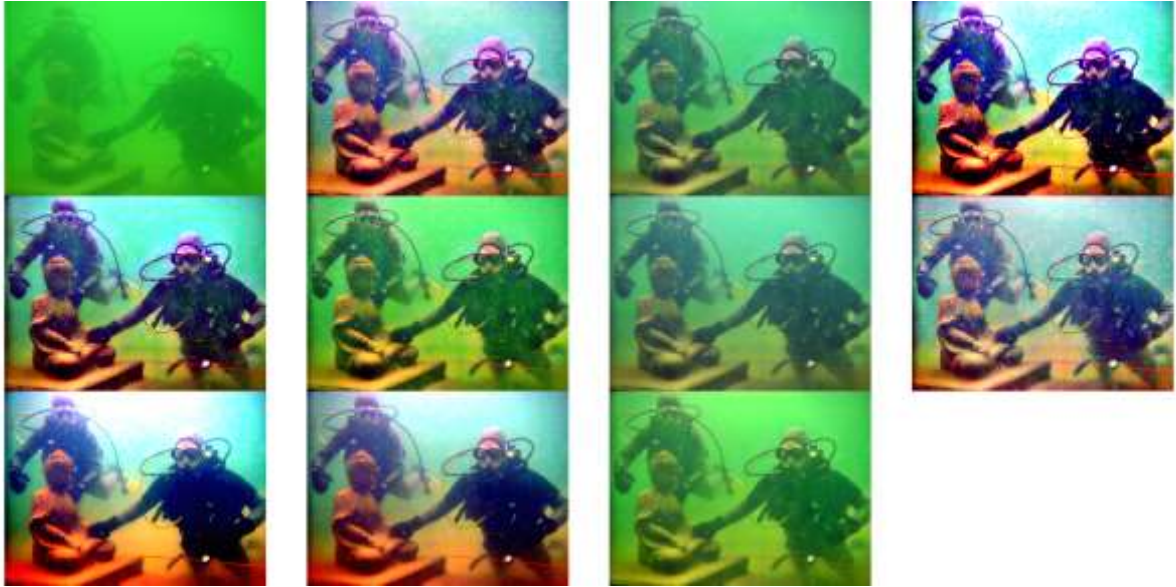

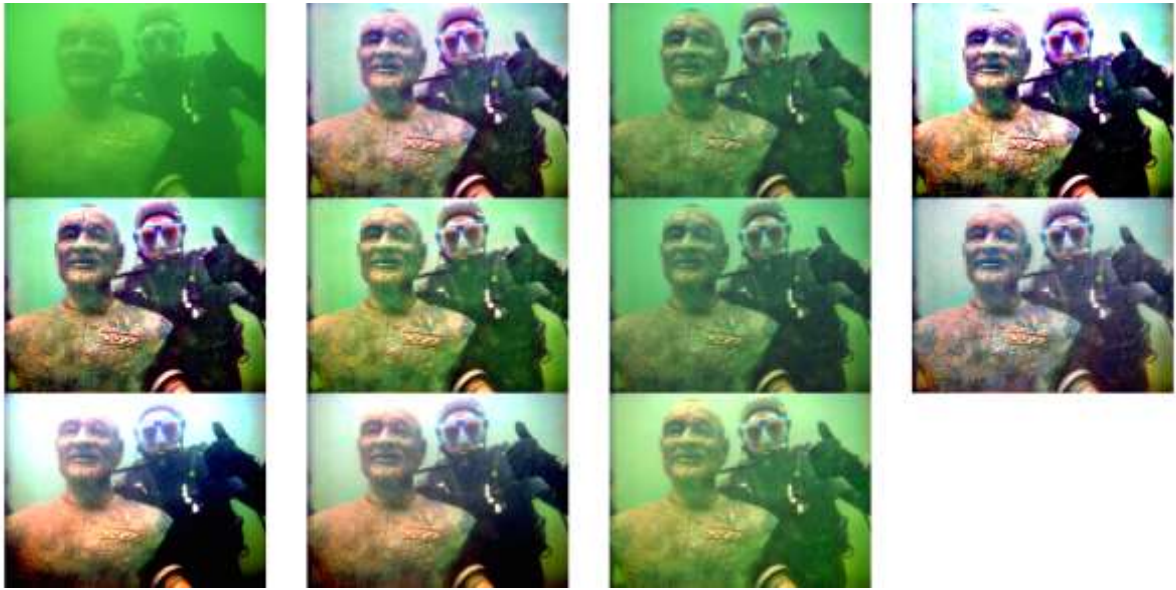

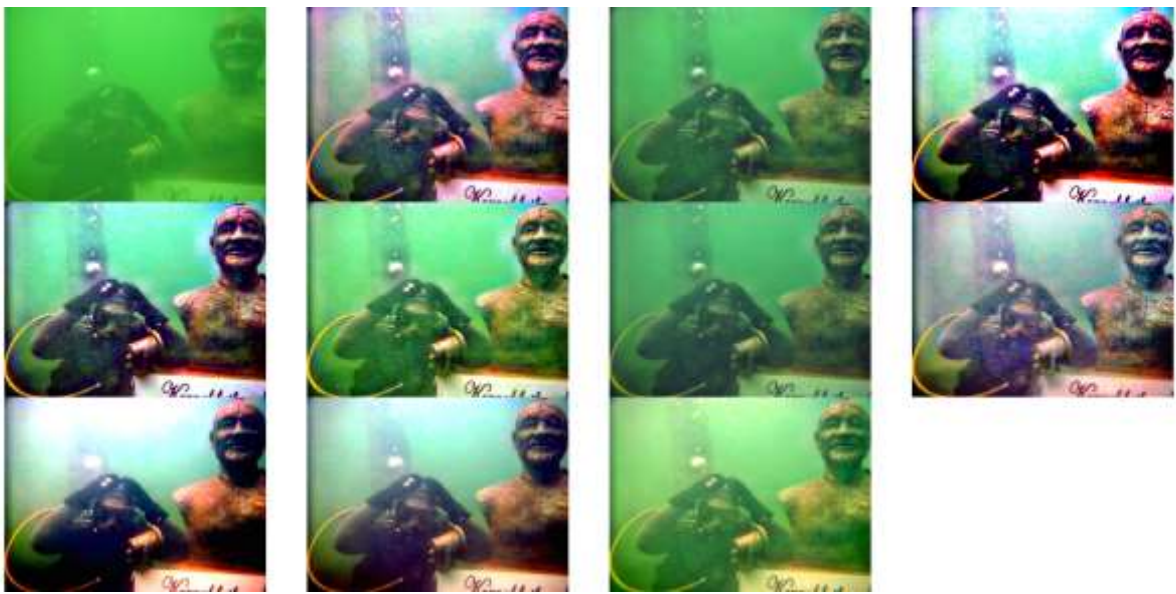

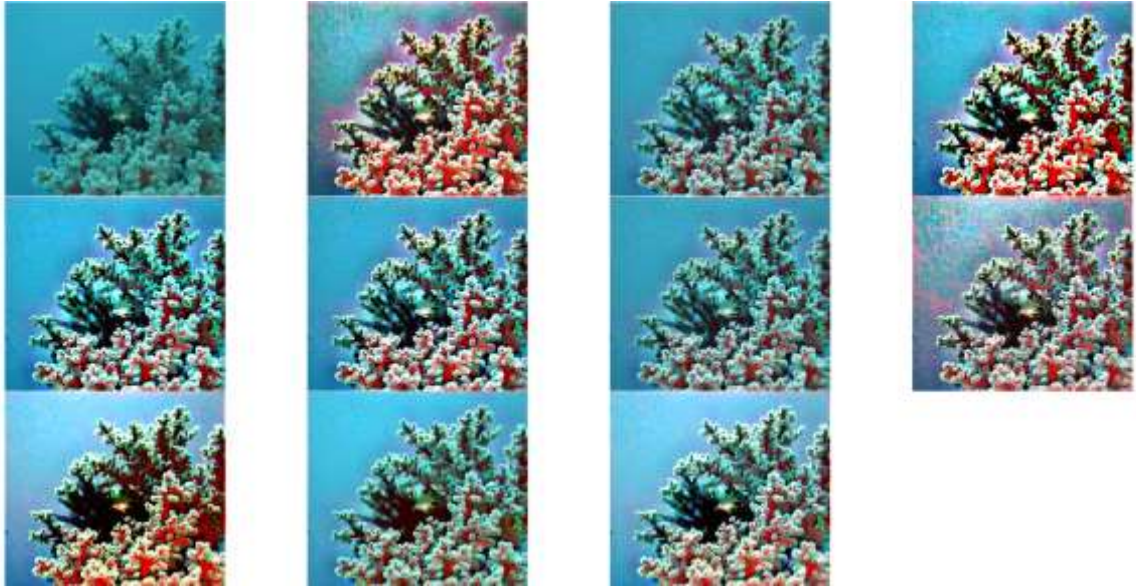

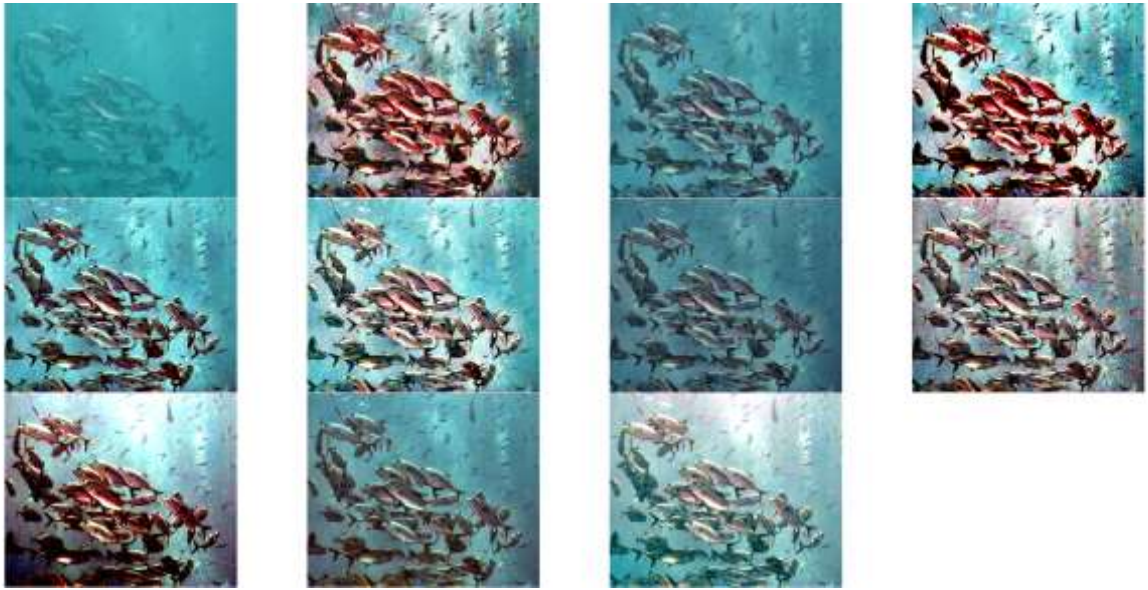

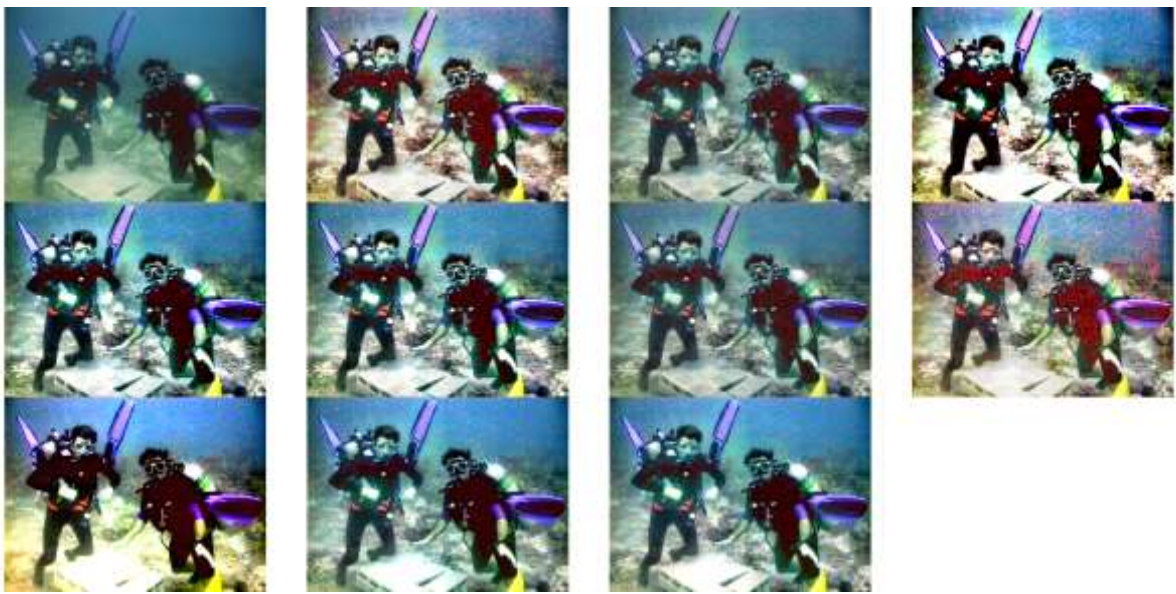

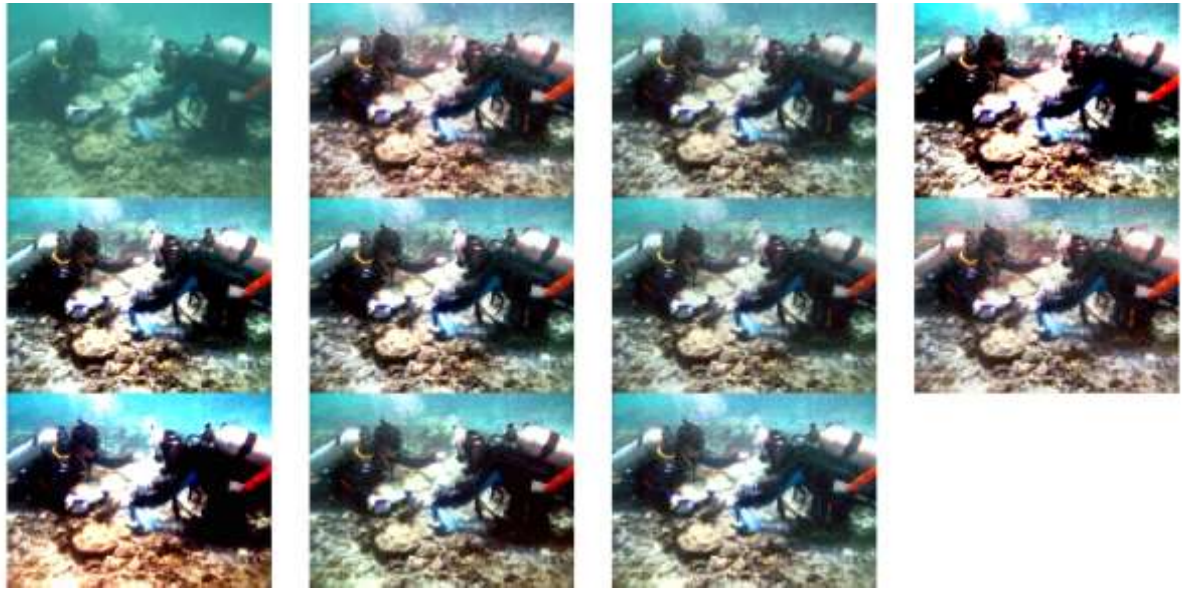

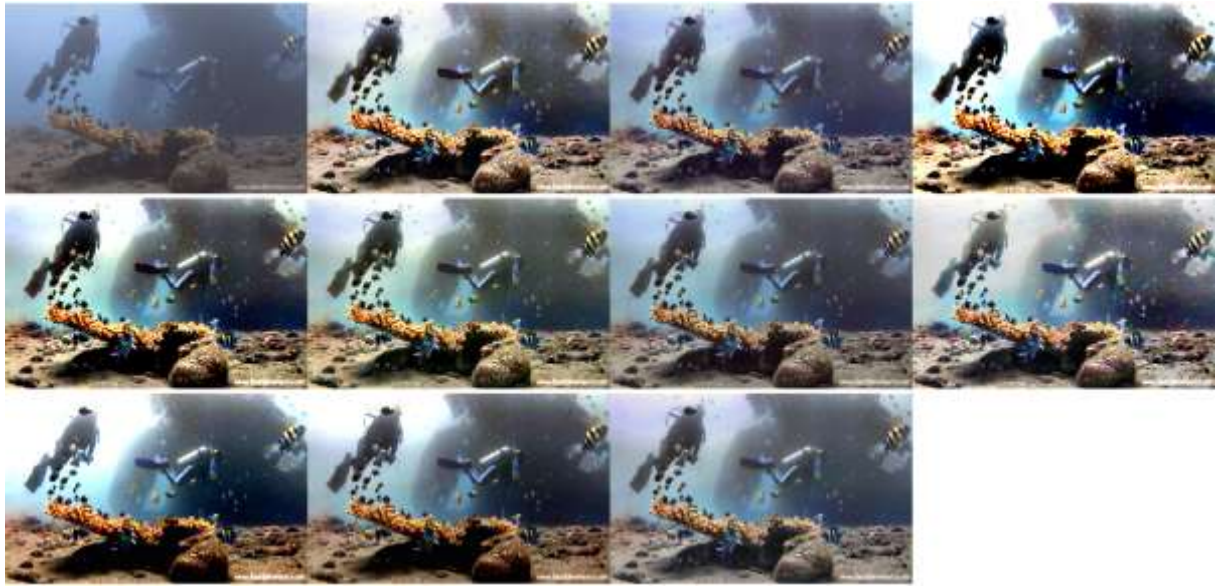

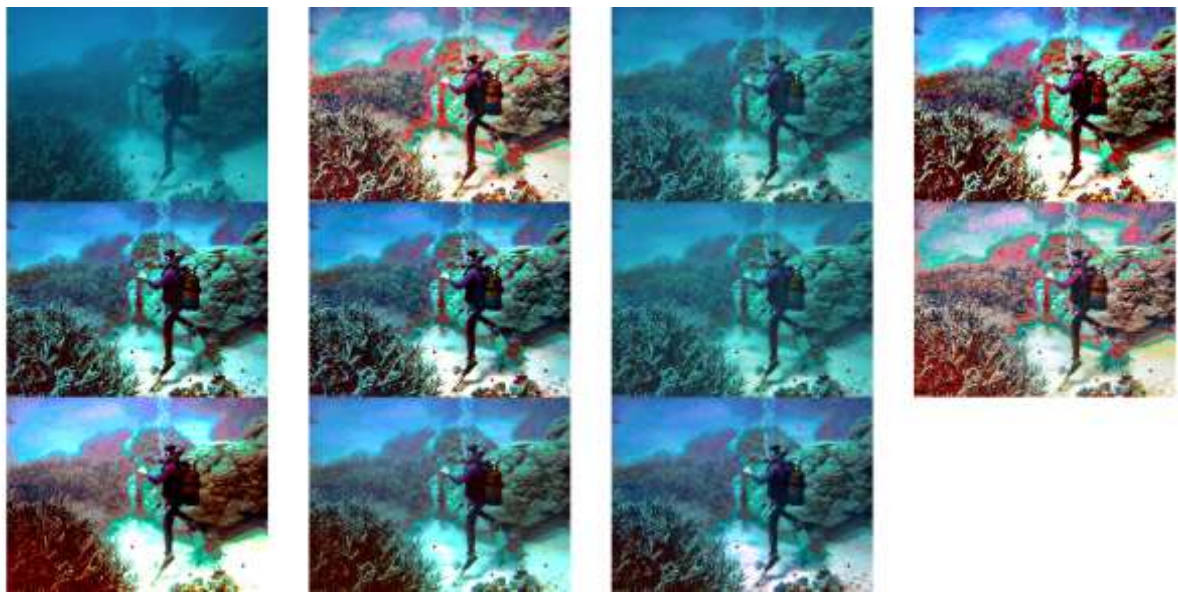

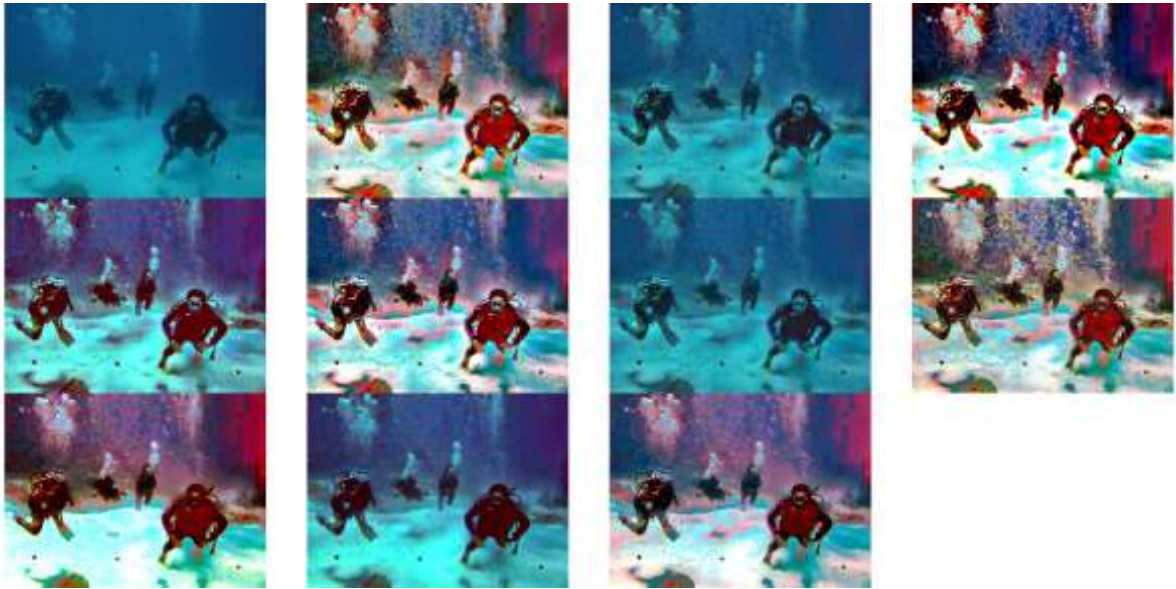

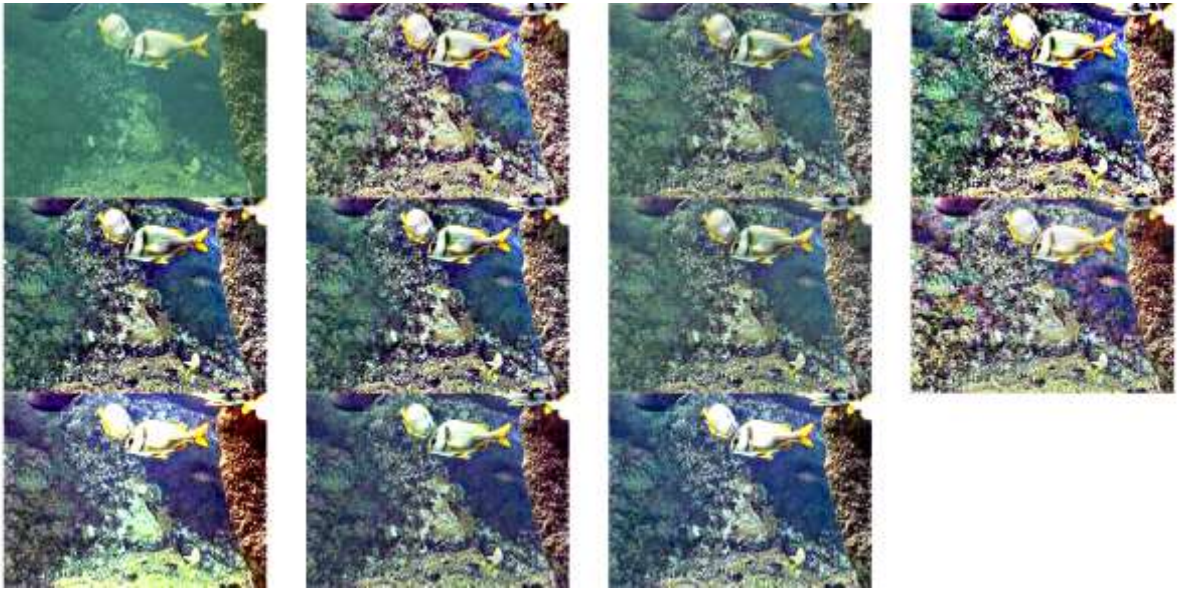

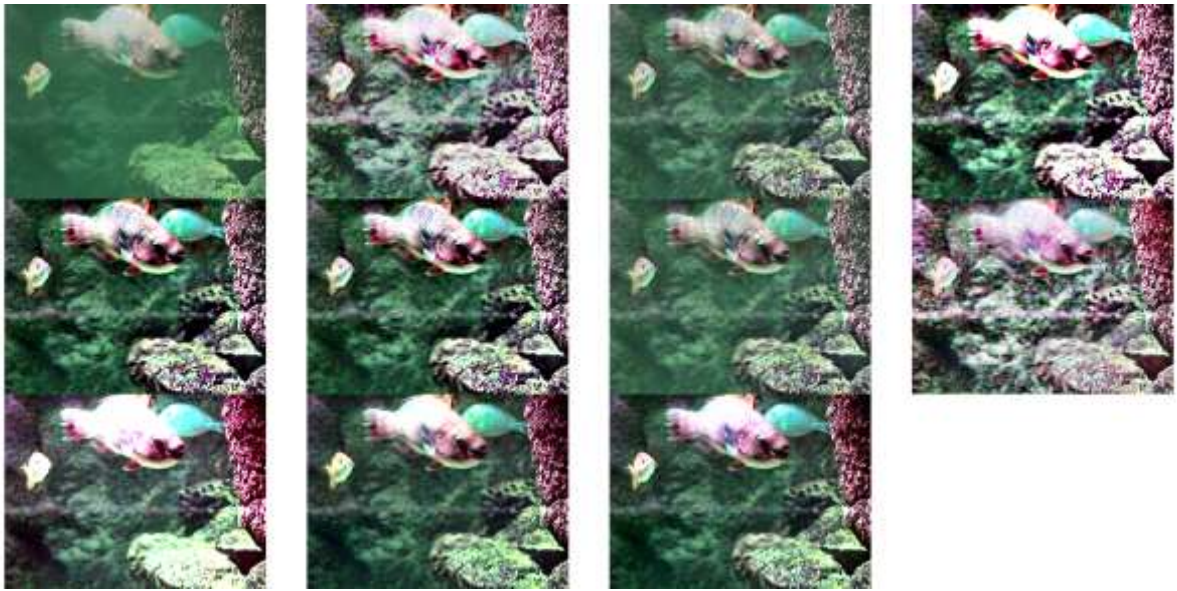

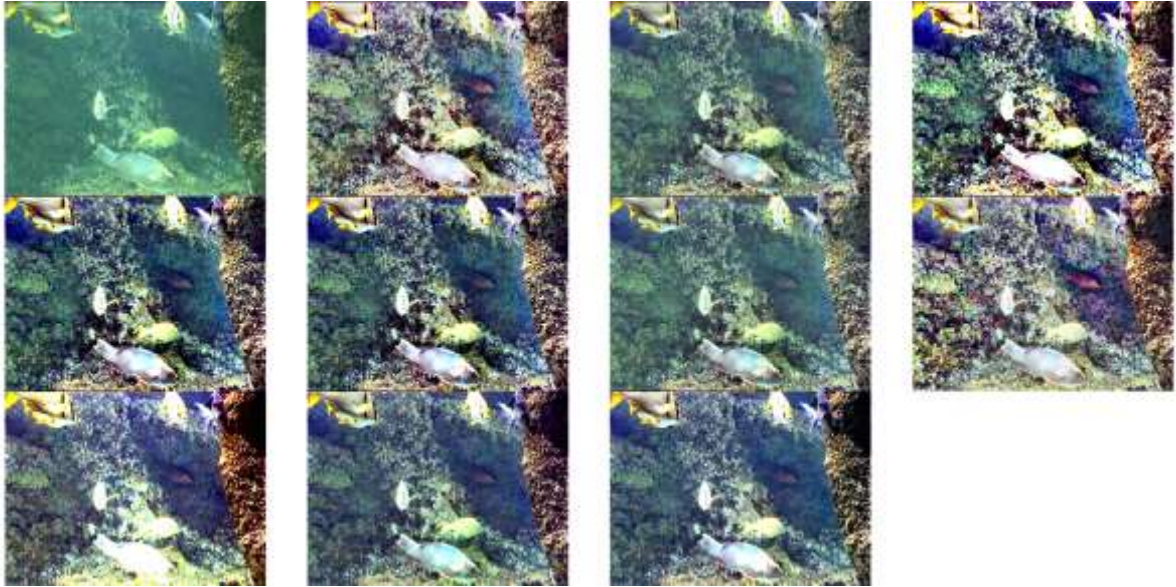

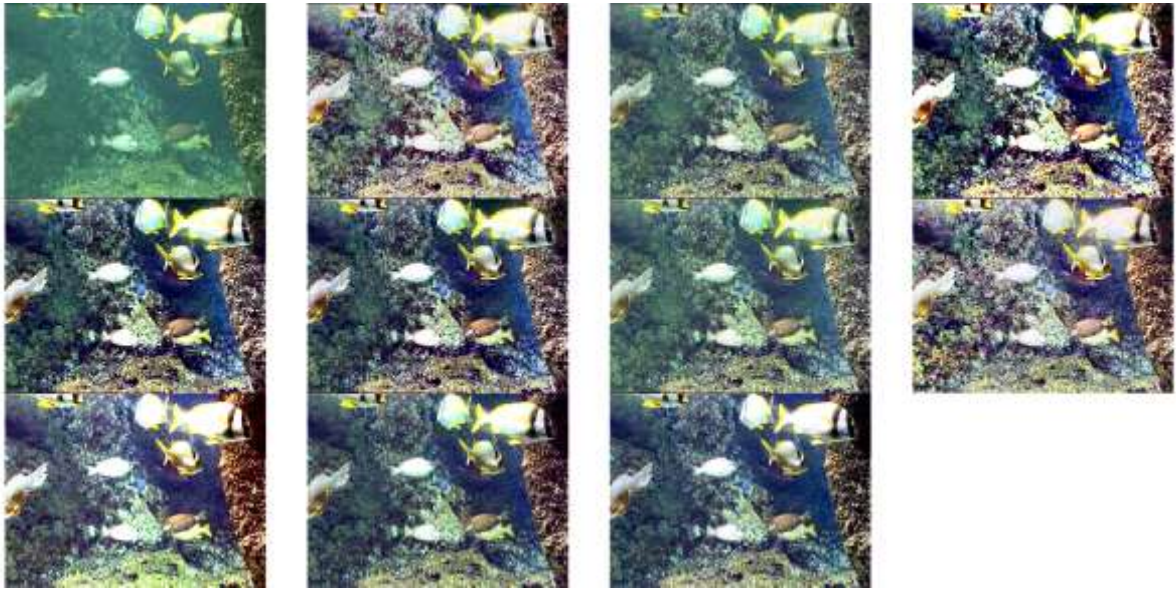

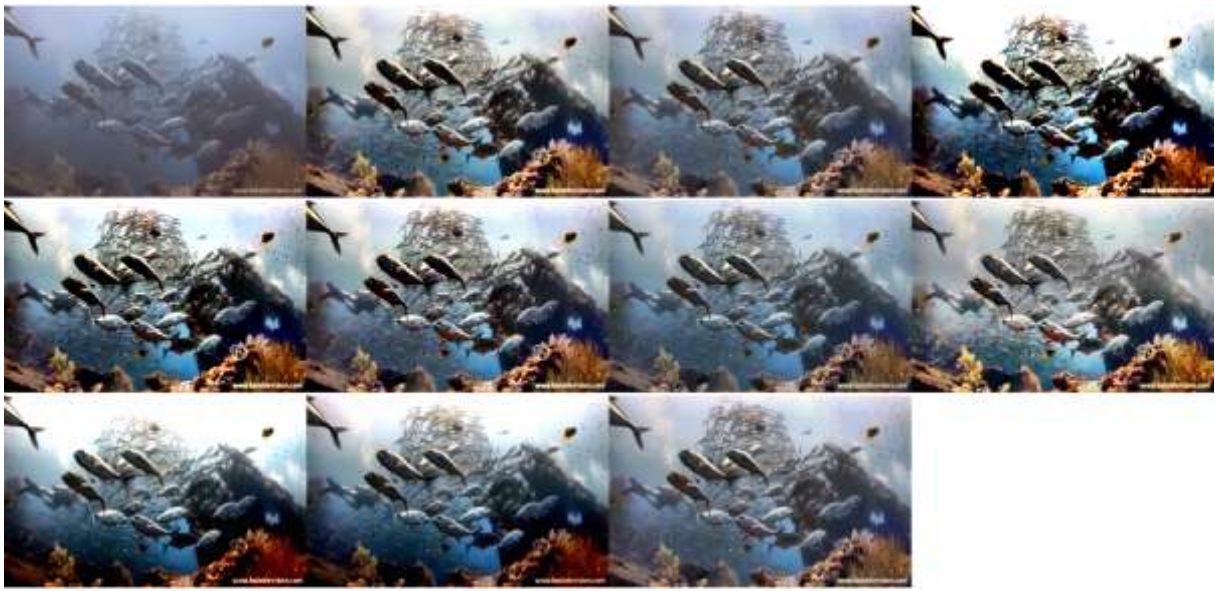

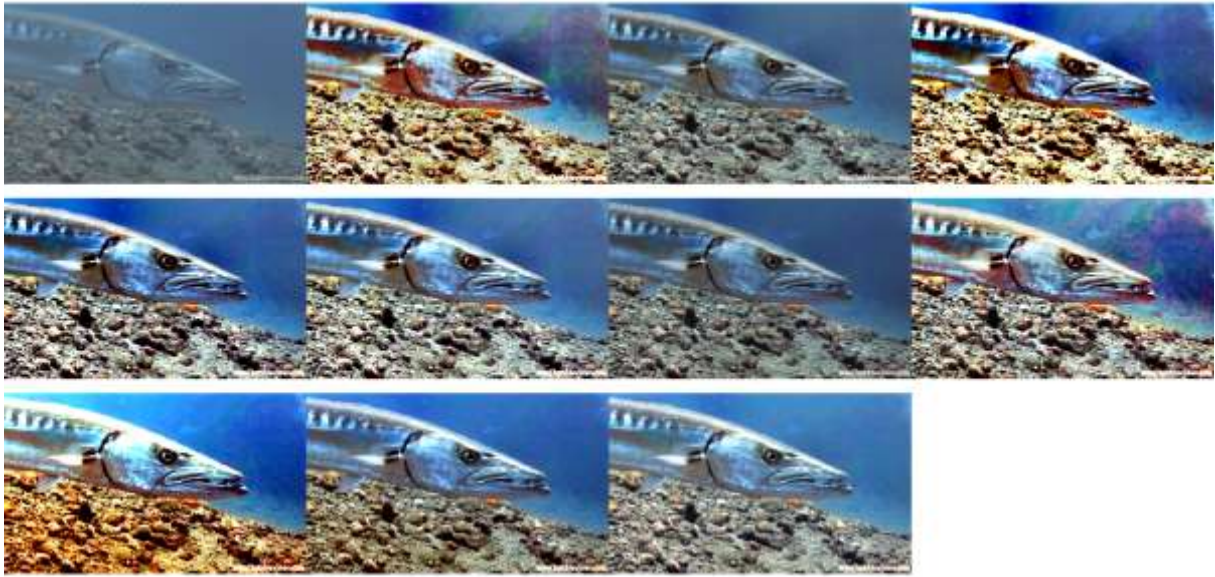

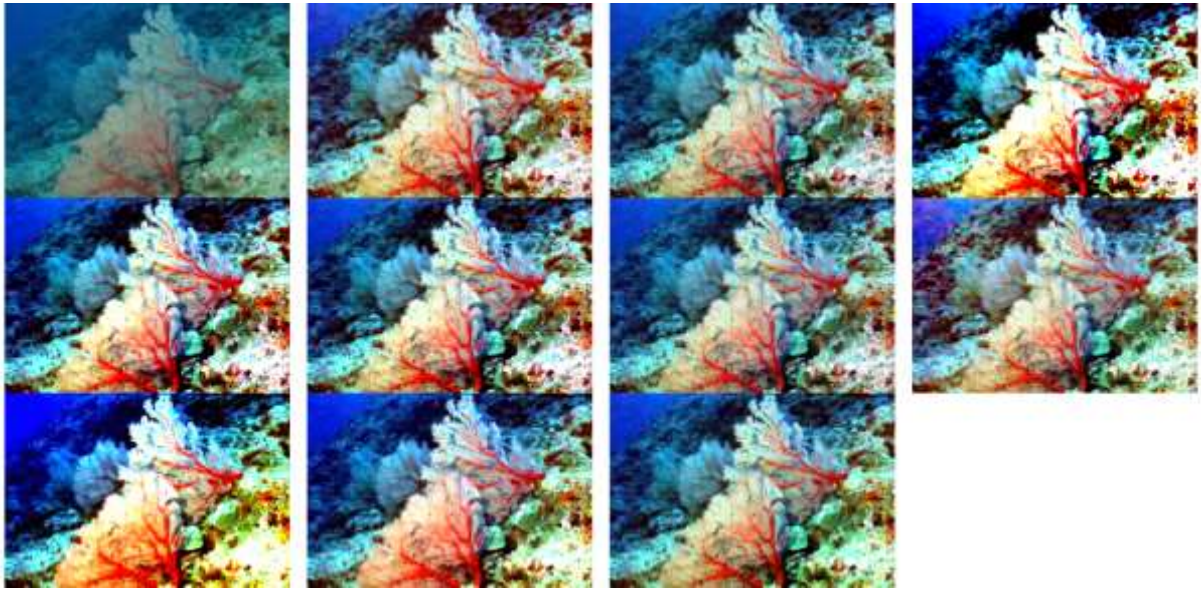

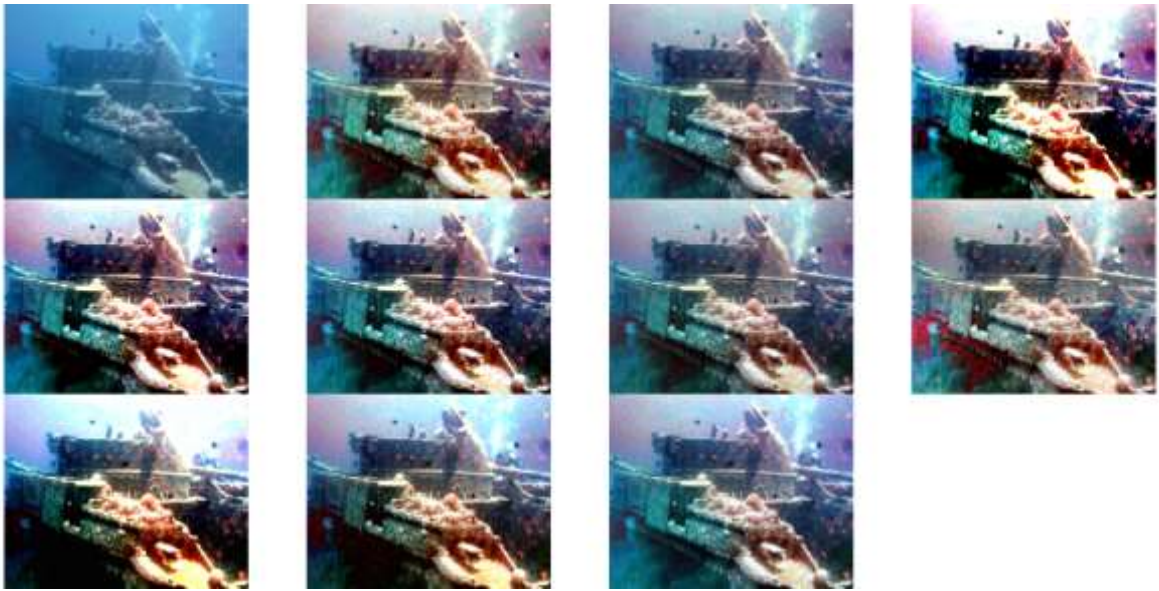

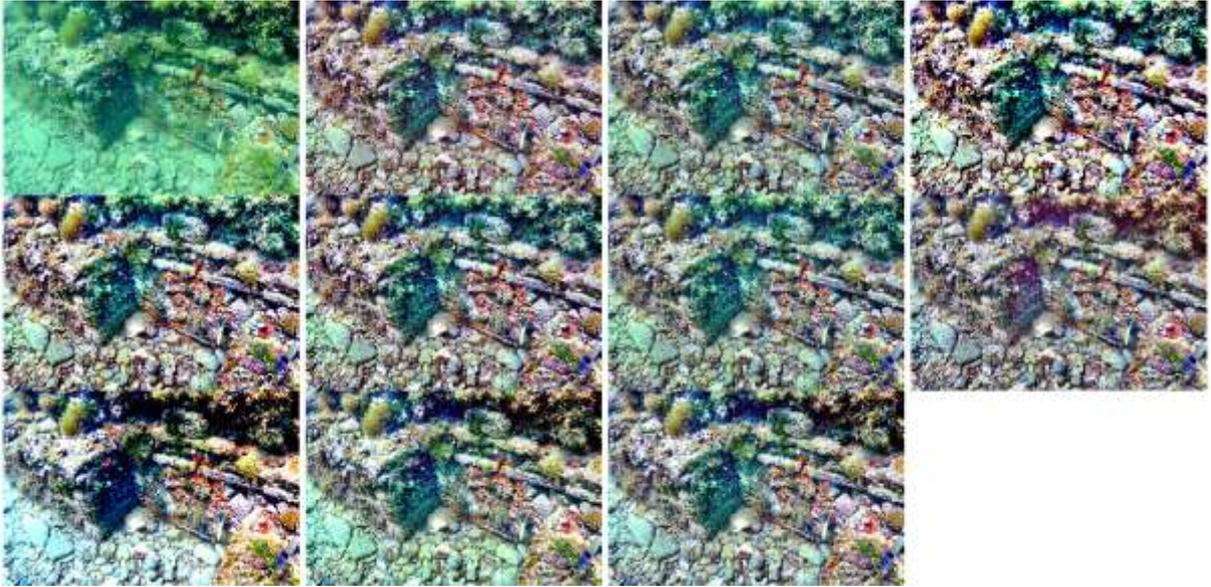

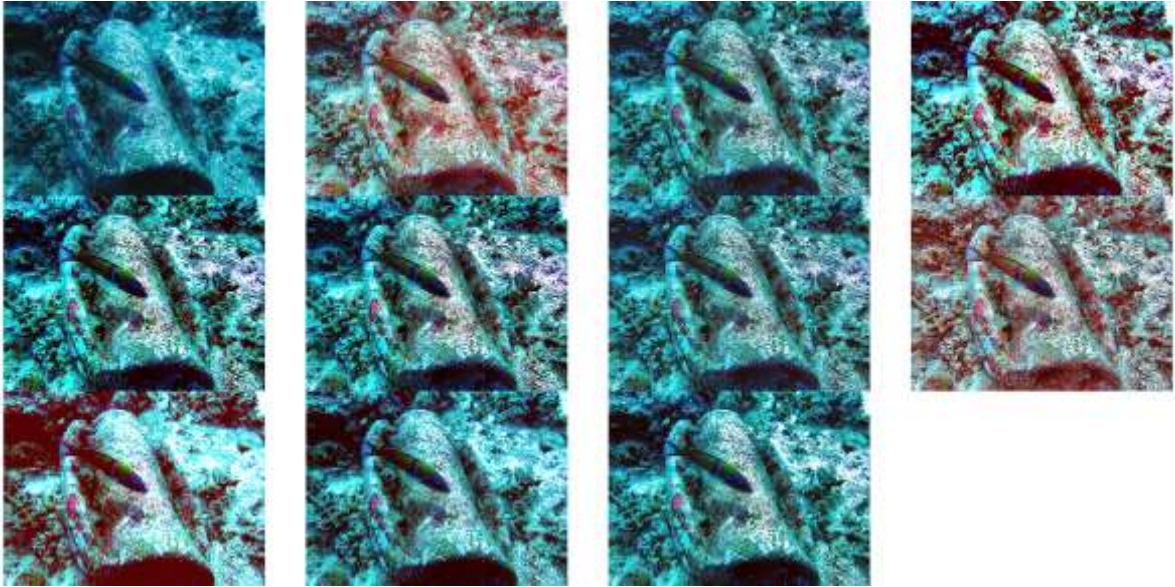

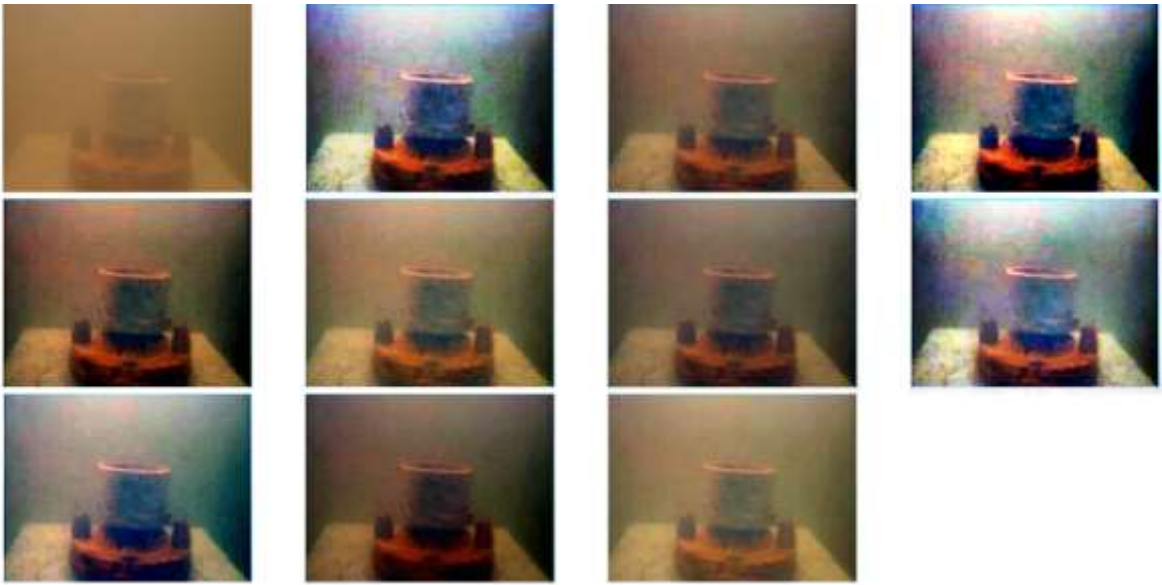

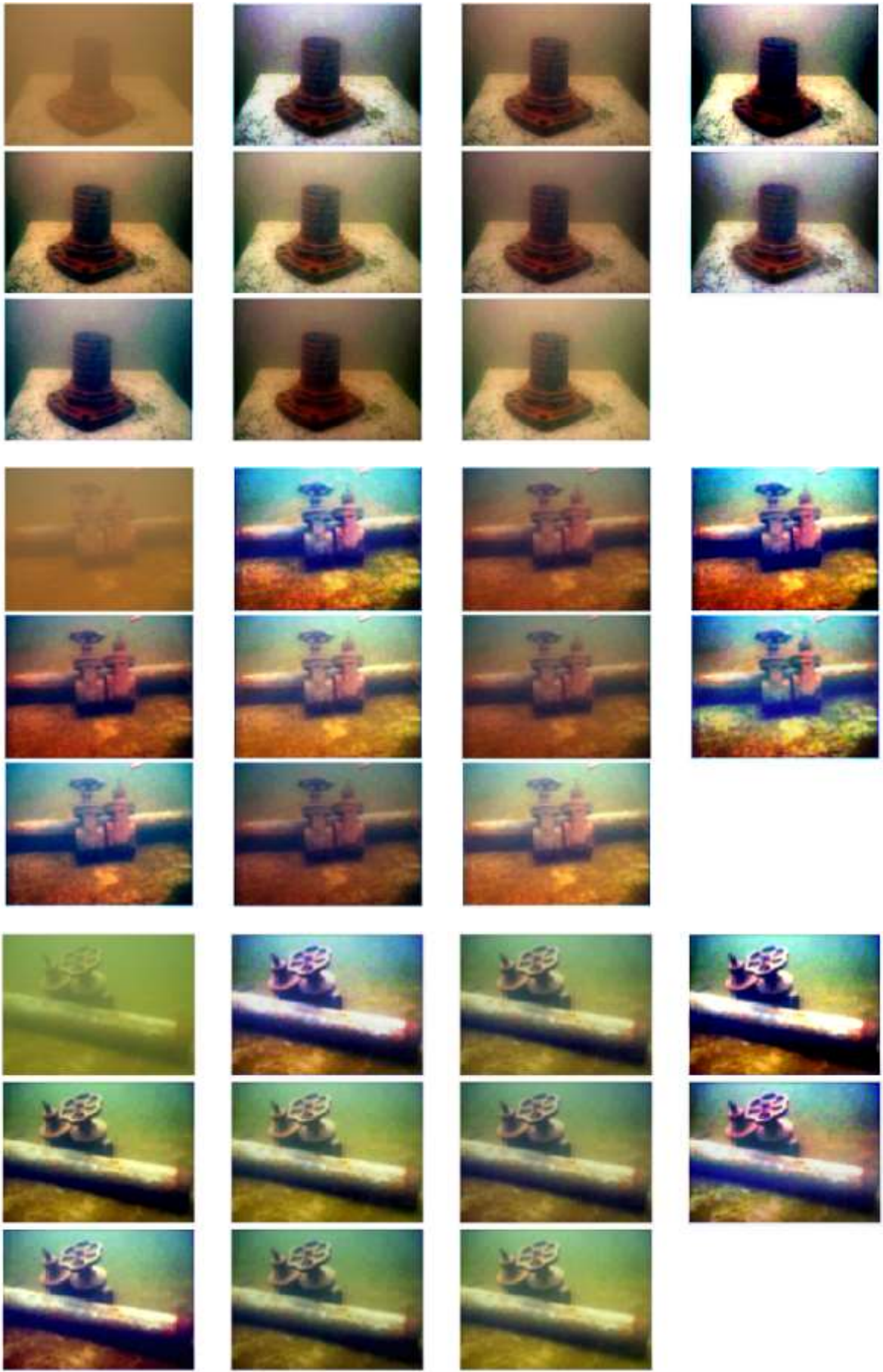

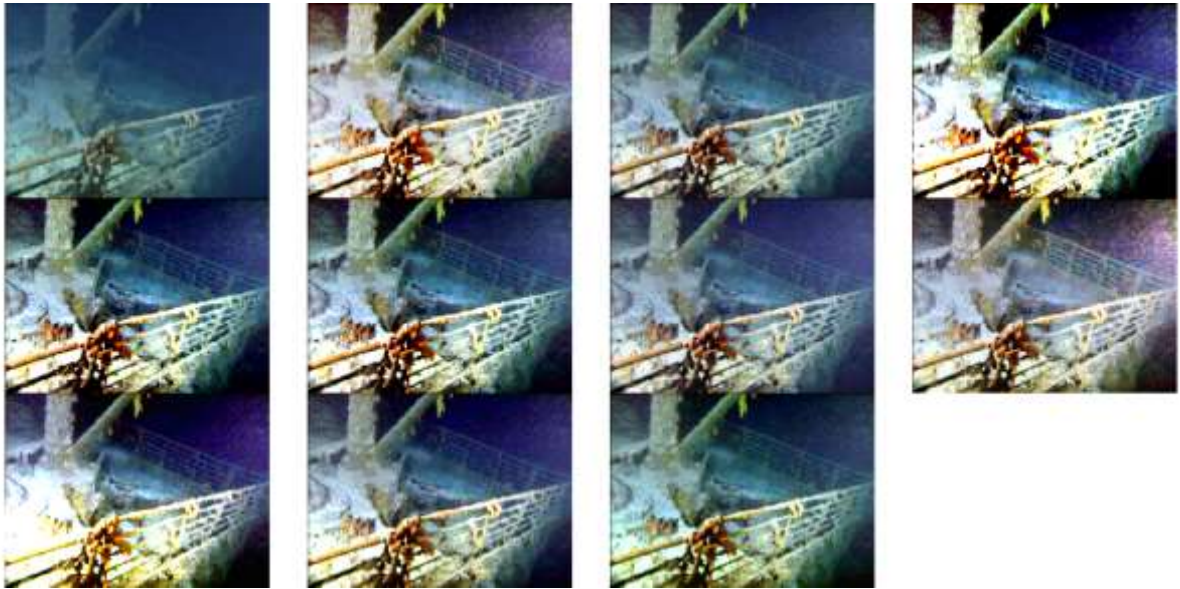

(a)

| Original | PDE-CLAHE-HS | PDE-CLAHE-GOC2 | PDE-CLAHE-GOC3 |
|---|---|---|---|
| PDE-CLAHE-PWL | PDE-CLAHE-CS | PDE-GOC2-CLAHE | PDE-HS-CLAHE |
| PDE-GOC3-CLAHE | PDE-PWL-CLAHE | PDE-CS-CLAHE | **KEY** |

(b)

**Fig. 2** (a) Images processed with various PDE-based configurations (b) key to figures

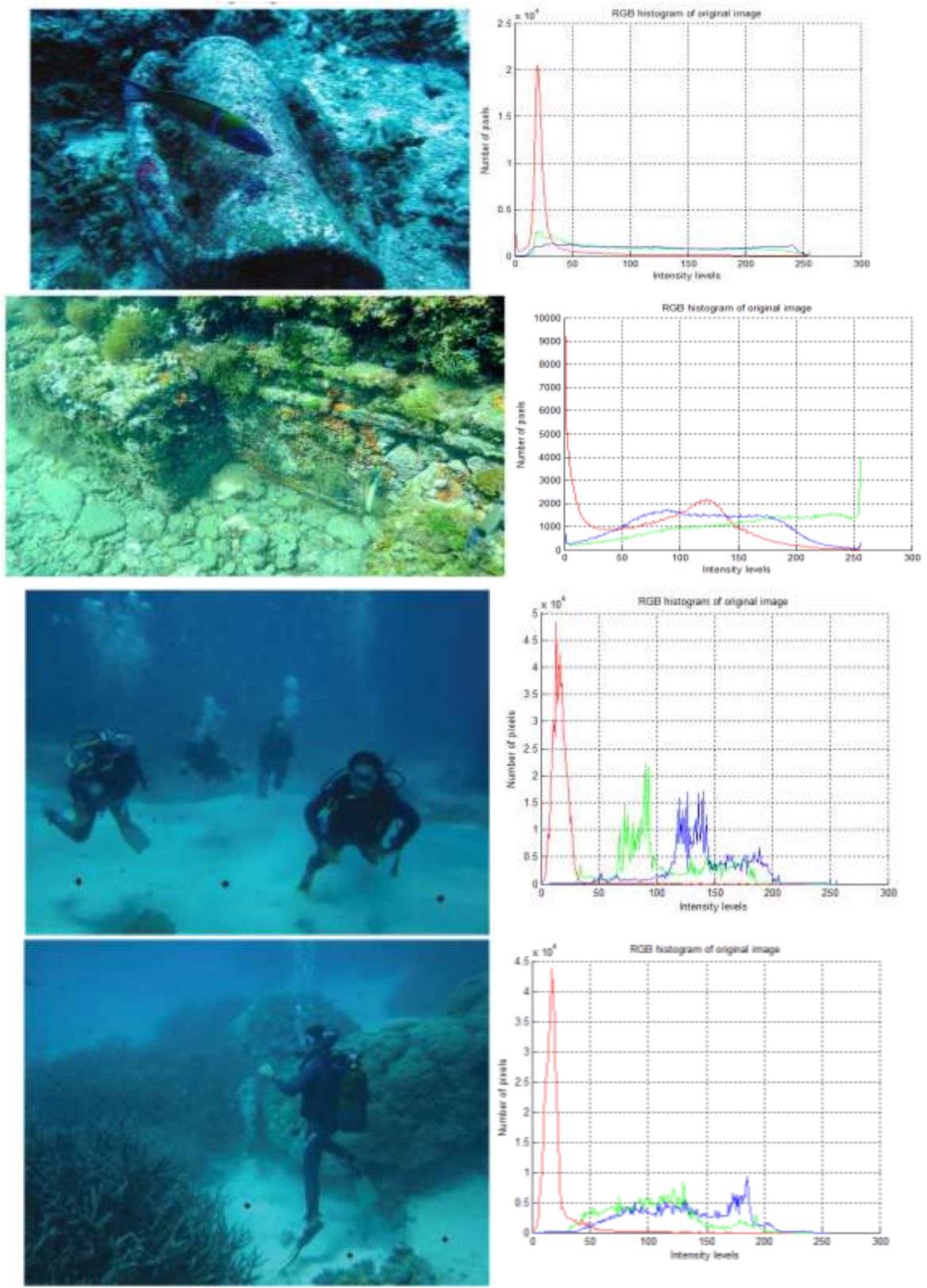

**Fig. 3** RGB colour histograms of original *Ocean jar* and *Ocean floor* images [6], *Divers and Diver* images [16]

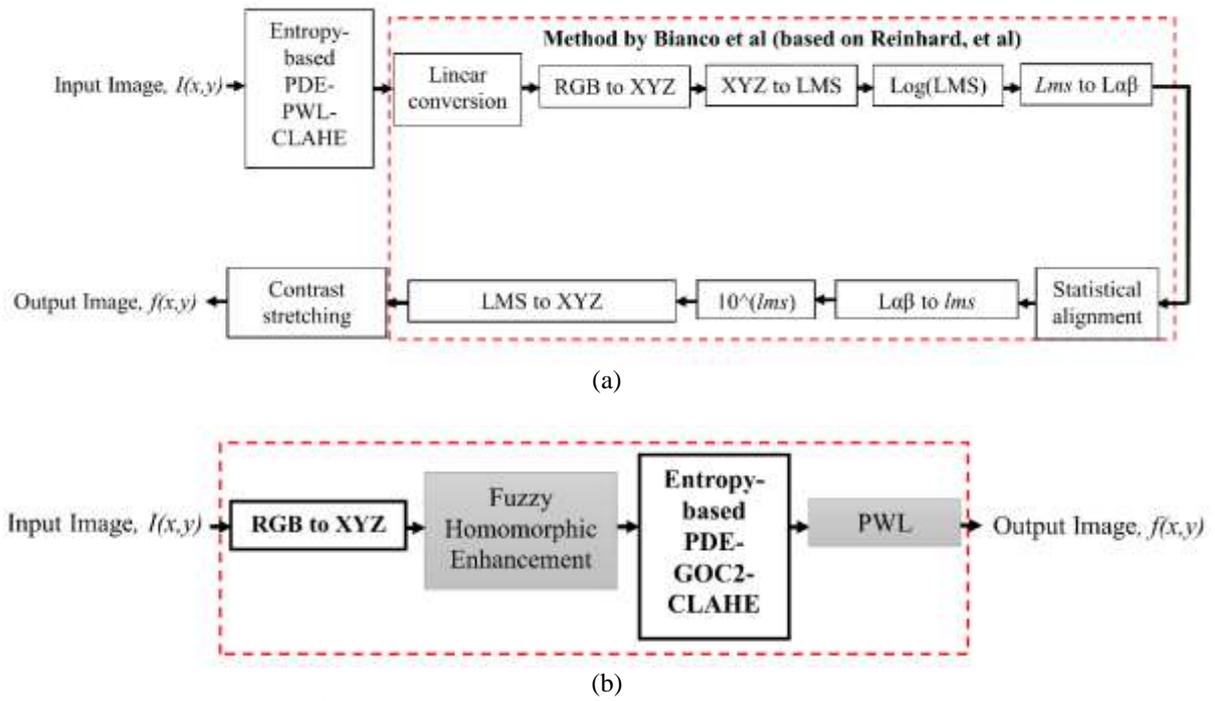

(a)

(b)

**Fig. 4** (a) PA-1 and (b) PA-2 for processing the images in Fig. 3

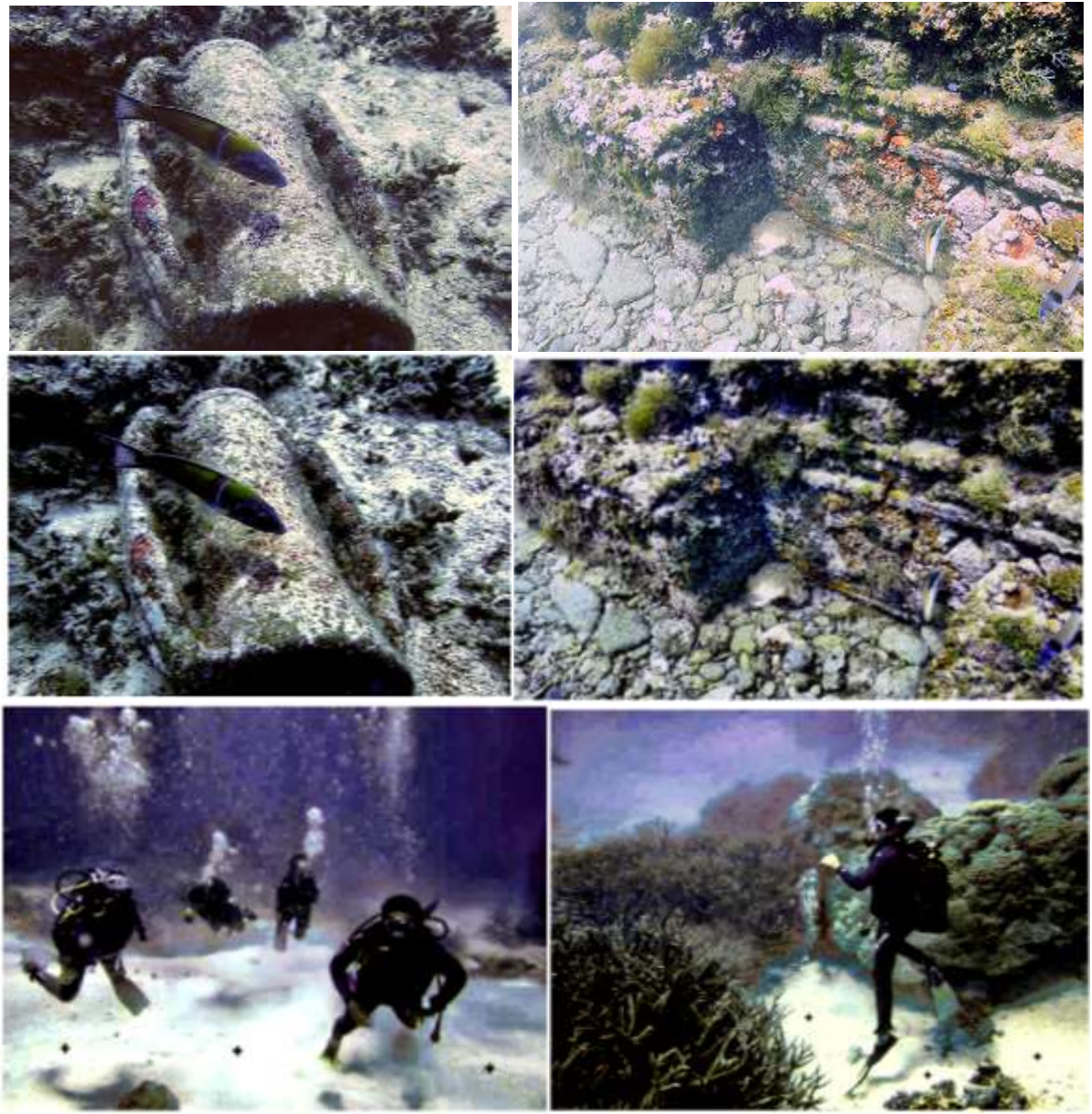

**Fig. 5** Results of Bianco, et al (first row) and PA-1 (second and third row)

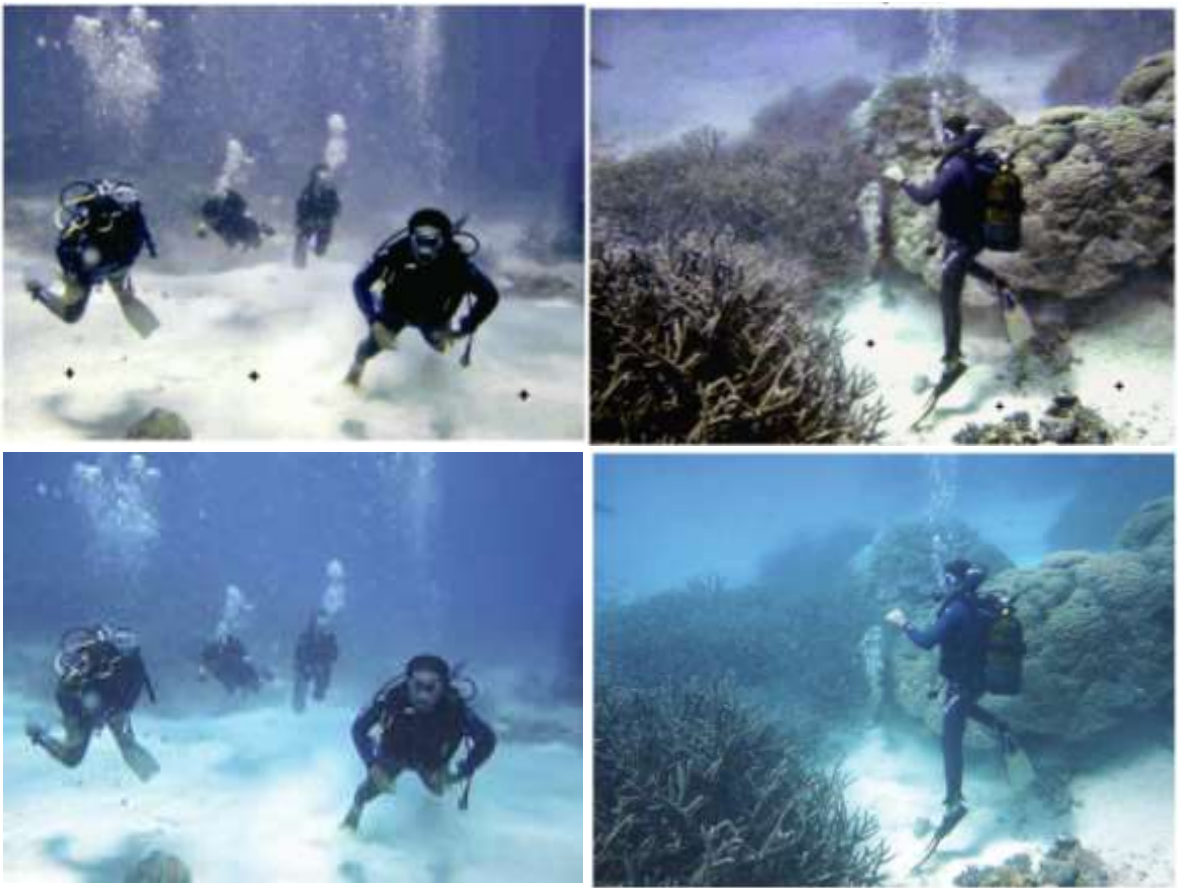

(a)                                                                    (b)

**Fig. 6** Results of PA-2 (first row) and CoLIP method by Gouinaud, et al (second row)